\documentclass[preprint,12pt]{elsarticle}

\usepackage{times} 
\usepackage[dvipsnames,table,svgnames]{xcolor}
\usepackage{graphicx}  
\usepackage{float} 
\usepackage{subfigure}
\usepackage{amsmath}
\usepackage{url}
\usepackage{multirow}
\usepackage{makecell, tabularx, threeparttable}
\usepackage{adjustbox}
\usepackage{xspace}
\usepackage{booktabs}
\usepackage{amssymb,mathrsfs,amsmath}
\usepackage{arydshln} 
\usepackage{amsfonts} 
\usepackage{algorithm}
\usepackage{algorithmic}
\usepackage{colortbl} 
\usepackage{xcolor} 
\usepackage[table]{xcolor}
\usepackage{pifont}

\usepackage{tikz}
\usepackage{hyperref}
\definecolor{hollywoodcerise}{RGB}{244, 0, 161}
\hypersetup{colorlinks,linkcolor={red},citecolor={hollywoodcerise},urlcolor={red}} 

\usepackage{amssymb}
\usepackage{amsmath}

\journal{Pattern Recognition}

\begin{document}

\begin{frontmatter}



\title{From Boundaries to Semantics: Prompt-Guided Multi-Task Learning for Petrographic Thin-section Segmentation} 




\affiliation[a]{organization={Research Institute of Petroleum Exploration and Development (RIPED)}}
\affiliation[b]{organization={The Hong Kong University of Science and Technology (Guangzhou)}}
\affiliation[c]{organization={The Hong Kong University of Science and Technology}}
\affiliation[d]{organization={Nanyang Technological University}}

\author[a]{Yili Ren}
\author[b]{Shiqi Wen}
\author[a]{Li Hou}
\author[b]{Dingwen Xiao}
\author[b]{Weiming Zhang}
\author[c]{Caleb Chen Cao}
\author[d]{Lin Wang}
\author[a]{Zilu Zheng}
\author[a]{Qianxiao Su}
\author[a]{Mingjun Zhao}
\author[b,c]{Lei Chen$^{*,}$}

\cortext[cor]{Corresponding author. \\
\textit{Email Address}: 
\texttt{renyili@petrochina.com.cn} (Yili Ren), \texttt{swen750@connect.hkust-gz.edu.cn} (Shiqi Wen),
\texttt{houli\_up@163.com} (Li Hou), 
\texttt{dxiaoaf@connect.hkust-gz.edu.cn} (Dingwen Xiao), 
\texttt{wzhang915@connect.hkust-gz.edu.cn} (Weiming Zhang),
\texttt{cao@ust.hk} (Caleb Chen Cao),
\texttt{linwang@ntu.edu.sg} (Lin Wang), 
\texttt{zhengzilu@petrochina.com.cn} (Zilu Zheng), 
\texttt{suqianxiao@petrochina.com.cn} (Qianxiao Su), 
\texttt{zmjdean@petrochina.com.cn} (Mingjun Zhao), 
\texttt{leichen@cse.ust.hk} (Lei Chen)}

\begin{abstract}

Grain-edge segmentation (GES) and lithology semantic segmentation (LSS) are two pivotal tasks for quantifying rock fabric and composition. However, these two tasks are often treated separately, and the segmentation quality is implausible albeit expensive, time-consuming, and expert-annotated datasets have been used.
Recently, foundation models, especially the Segment Anything Model (SAM), have demonstrated impressive robustness for boundary alignment. However, directly adapting SAM to joint GES and LSS is nontrivial due to 1) severe domain gap induced by extinction-dependent color variations and ultra-fine grain boundaries, and 2) lacking novel modules for joint learning on multi-angle petrographic image stacks.
In this paper, we propose \textbf{Petro-SAM}, a novel two-stage, multi-task framework that can achieve high-quality joint GES and LSS on petrographic images. 
Specifically, based on SAM, we introduce a Merge Block to integrate seven polarized views, effectively solving the extinction issue. Moreover, we introduce multi-scale feature fusion and color-entropy priors to refine the detection of ultra-fine grain boundaries. Finally, we leverage these high-quality edge priors to guide semantic segmentation, effectively extending the framework to the dual-task domain via joint optimization. To train Petro-SAM, we constructed the \textbf{first} multi-angle thin-section dataset suitable for both GES and LSS, uniquely enriching conventional edge annotations with lithology semantic labels.
Extensive evaluation on the public CPPID edge dataset and our dataset show that our Petro-SAM consistently outperforms traditional edge detectors, SAM-based baselines, and state-of-the-art semantic segmentation models. In particular, our Petro-SAM yields sharper grain boundaries and more accurate lithology segmentation maps.

\end{abstract}


\end{frontmatter}



\section{Introduction}
\label{Introduction}

Petrographic thin-section image understanding~\cite{zhang2024edge} is essential for quantitative rock characterization in sedimentology, structural geology, and reservoir analysis. Under cross-polarized light, mineral grains exhibit extinction behaviour, interference colours, and angle-dependent textural variations, making utilizing multi-angle polarized imaging 
a compelling strategy to provide complementary cues for two fundamental tasks: \textit{grain-edge segmentation (GES)} and \textit{lithology semantic segmentation (LSS)}.
GES relies on extinction-aligned boundary contrast for precise grain delineation, while LSS depends on polarization-induced colour shifts and birefringence patterns to differentiate mineral phases. 
Because both cues stem from the same polarization–mineral interaction, multi-angle imaging simultaneously captures grain-boundary contrast and mineral-discriminative signatures, naturally supporting unified GES–LSS modelling.
Despite their close physical relationship, \textbf{GES and LSS remain under-explored and are almost always treated independently}. Existing GES studies either rely on classical edge detectors or CNN-based methods such as HED and BDCN~\cite{xie2015holistically, he2019bi}, and more recently extinction-consistency networks designed specifically for multi-angle polarized images~\cite{zhang2024edge}. 
However, they focus exclusively on boundary detection. 
Conversely, LSS methods typically adapt semantic segmentation architectures such as U-Net and
HRNet~\cite{ronneberger2015unetconvolutionalnetworksbiomedical, wang2020deephighresolutionrepresentationlearning}, but often fail to capture fine-grained boundaries. Therefore, no existing framework jointly learns both GES and LSS from multi-angle polarized thin-section images. This gap leads to a natural yet unexplored problem: \textit{how to construct a multi-task framework in which GES and LSS jointly leverage multi-angle polarized cues to improve each other’s performance?}

A major challenge behind this gap is the lack of comprehensive datasets. High-quality petrographic annotation requires expert delineation of grain boundaries and mineral classes, making large-scale labeling costly and time-consuming. Existing datasets such as CPPID~\cite{zhang2024edge} cover only edge detection. To support multi-task learning, we construct a \textbf{new multi-angle petrographic dataset} (Sec .~\ref {sec: data construction}) containing \textbf{over 1,400 matched polarized sets} (7 views per set) with \textbf{fine-grained edge masks and four-class lithology annotations}. This dataset substantially broadens task diversity and provides a solid foundation for unified GES--LSS modelling.

\begin{figure}[t]
\centering
\includegraphics[width=\linewidth]{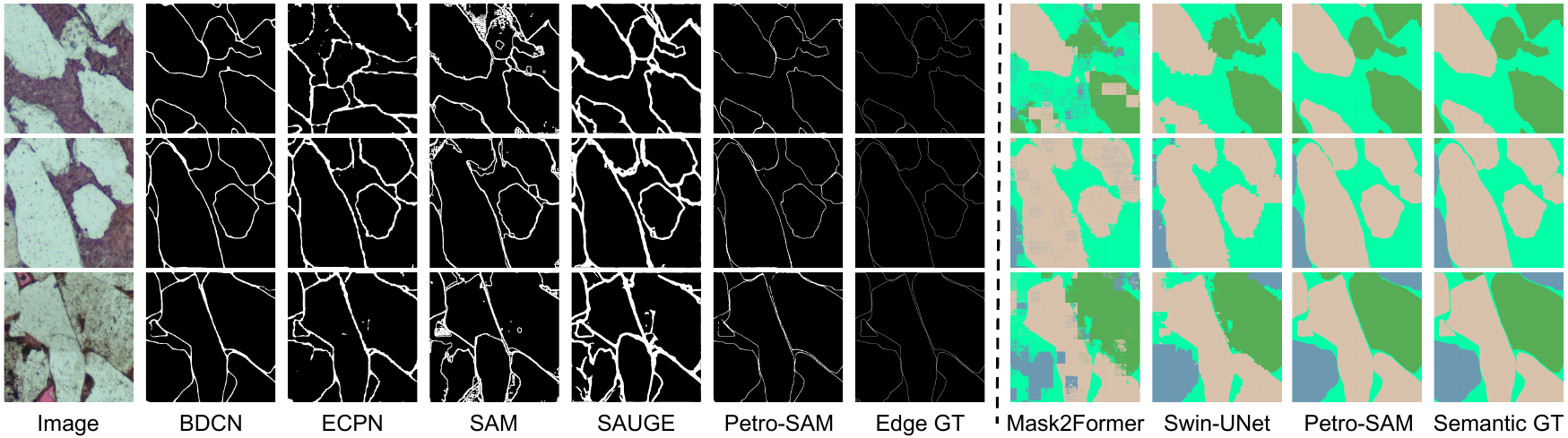}
\caption{Comparison of grain-edge segmentation (left) and lithology semantic segmentation (right) on multi-angle polarized thin-section images.}\
\label{fig: teaser}
\vspace{-0.5cm}
\end{figure}

Meanwhile, segmentation foundation models have rapidly advanced with the emergence of the Segment Anything Model (SAM)~\cite{kirillov2023segment},
which exhibits strong robustness in boundary alignment through its promptable segmentation paradigm.
Such capability makes SAM a promising starting point for grain-edge modelling.
However, directly adapting SAM to joint GES and LSS on petrographic thin-section images remains nontrivial.
First, a \textbf{severe domain gap} arises from extinction-dependent colour variation and ultra-fine grain boundaries,
which often leads to fragmented, over- or under-segmented masks when SAM is applied without adaptation.
Second, SAM \textbf{lacks a principled mechanism for joint learning from multi-angle polarized image stacks},
as it operates on single-view RGB inputs and produces instance-level masks without lithologic semantics.

To address these challenges, we propose \textbf{Petro-SAM}, a novel two-stage, multi-task framework for high-quality \emph{joint} GES and LSS on petrographic thin-section images. \textbf{Stage 1} (Sec.\ref{stage 1}) trains an edge-focused model built on SAM: a \emph{Merge Block} integrates seven polarized views via global attention to reduce extinction-induced fluctuations, while an \emph{Adaptation Block} aligns SAM features with petrographic textures. A \emph{Refine Block} then performs multi-scale fusion to capture ultra-fine, discontinuous contours, producing reliable grain-edge prompts. In \textbf{Stage 2} (Sec.\ref{stage 2}), the edge mask decoder employs an \emph{Entropy Block} to exploit cross-angle color consistency, and extends SAM’s prompt encoder to ingest the edge prior of Stage 1. Semantic mask decoder leverages the \emph{Adaptation Block}, which combines edge prior and semantic feature, and outputs a reliable LSS result. Boundary-aware and consistency supervision improve both contour fidelity and mineral-phase separability.

Extensive experiments on CPPID and our new multi-angle dataset show that Petro-SAM consistently surpasses traditional edge detectors, SAM-based baselines, and state-of-the-art semantic segmentation models, as shown in Figure \ref{fig: teaser}. Ablation studies verify the effectiveness of both stages, confirming that extinction-aware edge prompts and multi-task optimization jointly contribute to sharper grain boundaries and more accurate lithology segmentation maps. Together, these results demonstrate that Petro-SAM provides a unified and annotation-efficient solution for automated petrographic thin-section interpretation.

In summary, our contributions are \textbf{three-fold}: 
\begin{itemize}
\item We construct a new multi-angle petrographic thin-section dataset with over 1,400 matched polarized sets (seven views per set), providing fine-grained grain-edge masks and four-class lithology semantic annotations to enable unified evaluation of GES and LSS.
\item We propose \textbf{Petro-SAM}, a two-stage SAM-based framework for multi-angle thin-section analysis: \emph{Stage 1} fuses seven views via a \emph{Merge Block}, adapts features with an \emph{Adaptation Block}, and refines ultra-fine contours using a \emph{Refine Block} to generate edge prompts; \emph{Stage 2} strengthens GES with an \emph{Entropy Block} and a prompt-encoder extension, while using aligned features to support LSS.
\item Extensive experiments on diverse datasets demonstrate state-of-the-art performance for both GES and LSS tasks, with consistent gains over SAM-based baselines and strong semantic/edge models under polarization-driven domain shifts.
\end{itemize}

\section{Related Works}
\label{Related Works}
\subsection{Boundary Detection for Petrographic Images}
Petrographic boundary detection targets fine-grained features, but is hindered by complex textures, inconsistent contrast, and polarization-induced variations.
In the early stages, research primarily focused on traditional image processing methods, which utilized changes in pixel intensity or luminance to determine grain boundaries (e.g., edge-based methods~\cite{HEILBRONNER2000969}), or achieved segmentation by minimizing energy functions (energy-based methods~\cite{JUNGMANN201433}), as well as clustering similar pixels to generate tight borders (region-based methods~\cite{JIANG2018143}). Subsequently, the field entered a transitional phase where researchers began to adopt hand-crafted feature-based machine learning methods. They manually extracted features such as color and texture, and then employed data-supervised algorithms like artificial neural networks~\cite{FUETEN20071176} or k-means clustering~\cite{IZADI201538,Das2021Automatic} to predict segmentation maps. More recently, with the emergence of deep learning (e.g., VGGNet~\cite{simonyan2015deepconvolutionalnetworkslargescale}, ResNet~\cite{7780459}), the field underwent a paradigm shift toward end-to-end learning. Within this shift, HED~\cite{xie2015holistically} appeared as the first CNN-based edge detection network, followed by works such as RCF~\cite{liu2017richer}, BDCN~\cite{he2019bi}, PiDiNet~\cite{su2021pixel}, and SAUGE~\cite{liufu2025sauge}, which continued to improve performance by enhancing encoders and modifying backbone architectures.

 While these deep learning methods have shown effectiveness on images, they still face challenges in petrographic scenarios: ultra-fine grains, and polarization-induced domain shift remain hard to tackle, and high-resolution further introduce tiling artifacts; additionally, these approaches demand costly pixel-level ground truths and are highly sensitive to post-processing steps. \textit{To address these issues, we have constructed a relevant rock dataset, and therefore our method adopts a two-stage teacher–student scheme: a SAM-encoder-based teacher fuses multi-angle views and adapts to an edge-specialized network to produce high-quality edge prompts, which are then combined with color-entropy priors in the student for refined boundary extraction.
}

\subsection{Semantic Segmentation for Petrographic Images}
Semantic segmentation of petrographic images assigns a mineral label to each pixel, but faces challenges from multi-scale grain structures, inter-class visual similarity, and scarce annotations. The method lineage spans classical FCN~\cite{long2015fully} and encoder–decoder families ~\cite{ronneberger2015unetconvolutionalnetworksbiomedical} and atrous/ASPP designs ~\cite{chen2017rethinkingatrousconvolutionsemantic} for larger receptive fields, to multi-resolution context models~\cite{wang2020deephighresolutionrepresentationlearning}, and further to Transformer-based or hybrid segmenters~\cite{xie2021segformersimpleefficientdesign, REN2025548} capturing long-range dependencies. Specifically, FCN~\cite{long2015fully} established the foundation for end-to-end pixel classification; DeepLabV3~\cite{chen2017rethinkingatrousconvolutionsemantic} further enlarges the receptive field via atrous convolutions and ASPP; CCNet~\cite{huang2019ccnet} captures non-local context with criss-cross attention; Swin-UNet~\cite{cao2022swin} couples hierarchical Transformers with a U-shaped decoder to achieve global–local balance; and Mask2Former~\cite{cheng2021mask2former} unifies the segmentation process with a query-based mask framework, improving cross-class consistency. Additionally, domain adaptation and low-shot strategies (style transfer~\cite{gong2019dlowdomainflowadaptation}, adversarial adaptation~\cite{tsai2019domainadaptationstructuredoutput}, self-training~\cite{zhang2021prototypicalpseudolabeldenoising}, prototypical learning~\cite{xu2022semisupervisedsemanticsegmentationprototypebased}) mitigate microscope-condition shifts and data scarcity, while post-processing and priors (CRF~\cite{krähenbühl2012efficientinferencefullyconnected}, morphology~\cite{serra1983image}, topology/boundary-aware losses~\cite{kervadec2019boundary}) refine grain borders and small objects. 

Despite the aforementioned progress in mineral-phase recognition, inter-class confusion and intra-class variability persist; tiny and slender phases are easily missed; high-resolution whole-slide inference is memory-intensive; purely semantic supervision often blurs boundaries.
To improve data efficiency and cross-domain transfer, promptable general segmentation paradigms have gained traction. \textit{Thus, our second stage introduces a multi-task decoder: conditioned on the stage-1 edge prompts and multi-angle features, we jointly optimize boundary detection and lithology segmentation, enhancing grain-boundary alignment while preserving regional consistency.}

\subsection{Segment Anything Model (SAM)}
Segment Anything Model (SAM) has revolutionized the image segmentation paradigm by reformulating the task into a prompt-driven general problem, achieving exceptional Zero-Shot transfer capabilities and strong boundary alignment due to its large-scale data pre-training. To adapt SAM to specific domains, various strategies have been explored, including lightweight fine-tuning via LoRA or Adapters~\cite{chen2023sam,lai2024lisa}, tackling prompt dependency through Prompt Engineering~\cite{piater2025prompt}\ or cascaded prompting~\cite{ye2023sa}, and using knowledge distillation~\cite{zhou2025edgesam} to transfer its rich feature representations to more efficient models. In rock image analysis, SAM has been applied to rock fragment segmentation \cite{zhao2024identification}, mineral grain identification \cite{zhang5158450few}, and utilized for digital rock analysis with edge semantics \cite{WANG2025100292}. 

However, applying SAM or its variants to complex rock thin-section analysis faces inherent limitations: 
(1) Masks are often over-smoothed or leak, struggling to capture low-contrast mineral grain boundaries, and the auto-masking mode frequently leads to over-segmentation or a granularity mismatch \cite{zhang5158450few}; (2) SAM lacks semantic information, as its mask output cannot directly perform lithology identification \cite{zhao2024identification}; and (3) Existing methods, primarily based on single images, ignore critical domain priors, such as the essential multi-angle polarized light information 
and local color variations. 
\textit{To address these challenges, we propose Petro-SAM. Our model uses the SAM Vision Encoder as a transferable feature backbone within a two-stage multi-task framework. }

\subsection{Multi-task Learning}
Multi-Task Learning (MTL) is a powerful paradigm that jointly trains related tasks (e.g., boundary detection, semantic segmentation) on shared representations to leverage complementary supervision, thereby improving model generalization and stability \cite{ruder2017overview}. The evolution of MTL has progressed from simple hard parameter sharing to sophisticated soft sharing mechanisms (e.g., Sluice \cite{ruder2017sluice}, MTAN\cite{liu2019end} MTLoRA \cite{agiza2024mtlora}), and advanced optimization techniques, such as ConsMTL \cite{qin2025towards} and CoNAL \cite{yue2023learning}, have emerged to effectively mitigate the long-standing problems of gradient conflict and negative transfer. In segmentation, MTL is particularly effective at coupling boundary guidance with region refinement to simultaneously boost local precision and regional consistency \cite{wang2020boundary} by utilizing structural priors \cite{caliva2019distance}.

However, applying MTL to fine-grained petrographic image analysis and integrating large model priors presents critical challenges: Traditional MTL frameworks struggle to effectively integrate the general priors from large models (like SAM) with domain-specific structural priors (e.g., color boundaries), and optimization conflicts remain detrimental to the high-precision boundary localization and accurate semantic classification required for petrography. \textit{To address these challenges, we propose the MTL-based Petro-SAM framework. This framework adopts a design with a shared backbone and dual heads, employing specific mechanisms to manage conflicts. This "Prompt + Multi-Task" design fosters the co-evolution of boundary and regional representations, achieving both precise boundary modeling and robust lithology recognition in petrographic images.}

\begin{figure}[t]
\centering
\includegraphics[width=\linewidth]{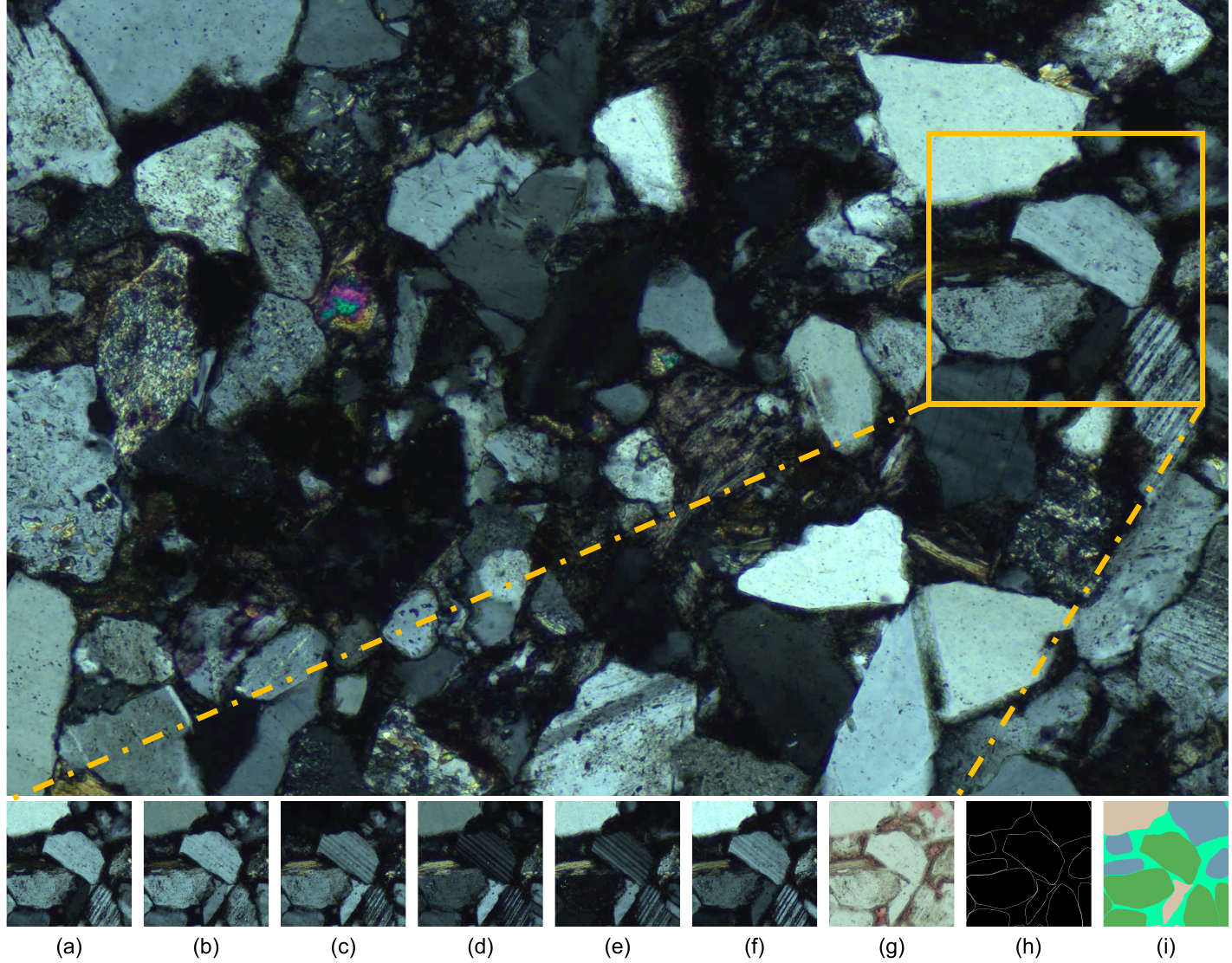}
\caption{Dataset visualization of multi-angle polarized thin-section samples with seven cross-polarized views (a–g), and corresponding grain-edge (h) and lithology semantic (i) annotations.}
\label{fig: dataset}
\end{figure}

\begin{figure}[t]
\centering
\includegraphics[width=\linewidth]{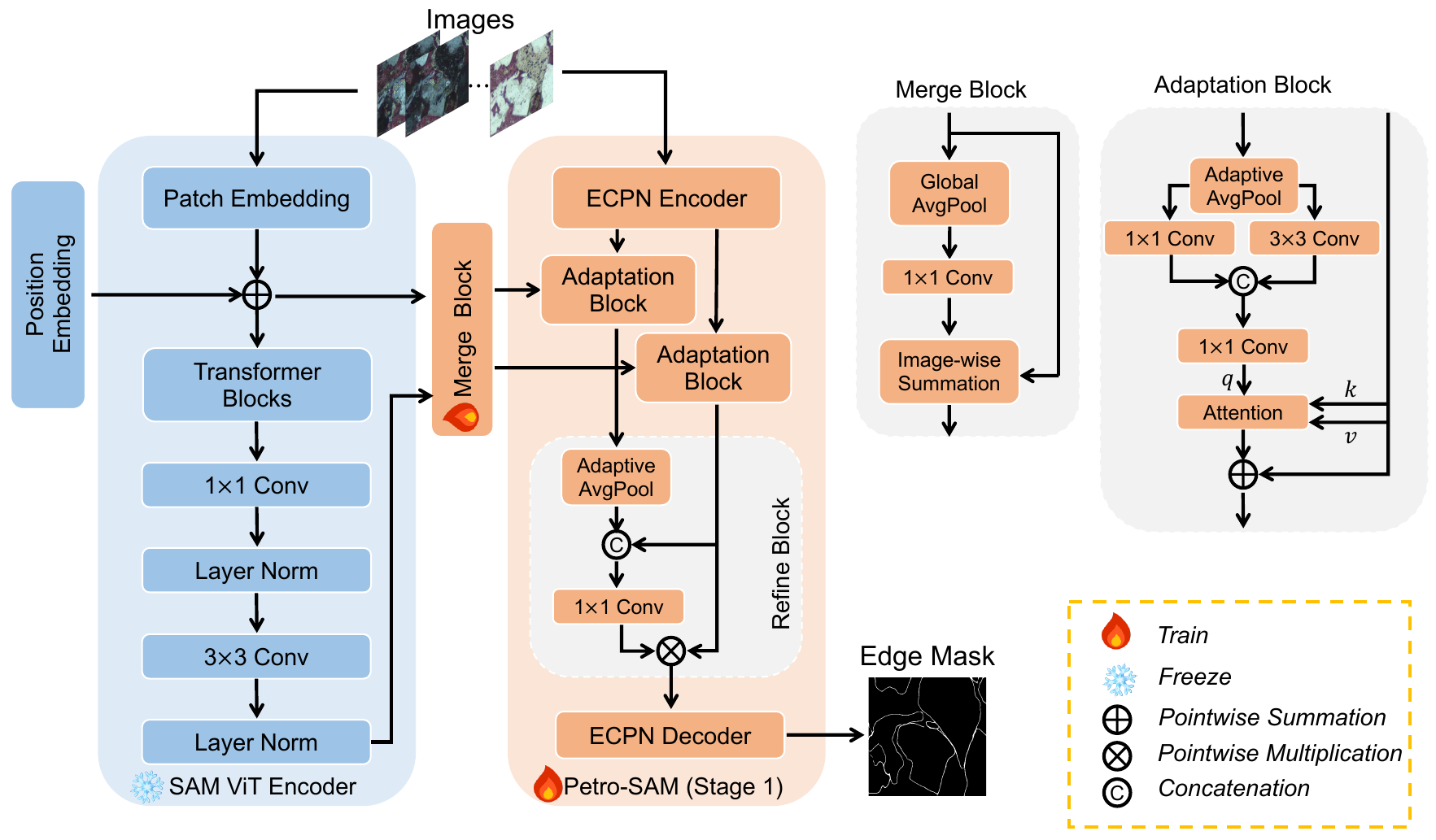}
\caption{Visualization of Petro-SAM's Stage 1 framework, Refine Block, Merge Block, and Adaptation Block.}\
\label{fig: stage_1}
\vspace{-0.5cm}
\end{figure}

\section{Methodology}
\label{Methodology}
\subsection{Our Dataset Construction}
\label{sec: data construction}
During our research, we identified a critical limitation: no single existing dataset simultaneously caters to both grain edge detection and lithological recognition tasks in rock thin-section images. To address this gap and the need for high-precision, multi-dimensional analysis, we constructed a new dataset consistent in design with the public CPPID (Cross-polarized Petrographic Image Dataset)~\cite{zhang2024edge}. While CPPID focuses solely on edge delineation, our dataset introduces a key extension: we incorporated semantic lithology labels alongside the pixel-level edge annotations. These semantic categories cover quartz, feldspar, debris, and background. Of the 2,000 samples initially collected, 1,400 high-quality image groups were precisely retained, divided into 1,120 for training and 280 for testing. This dual-label approach is crucial for supporting the joint learning of edge extraction and multi-class semantic segmentation within our framework.

Data preparation and acquisition strictly followed relevant industry and group standards (e.g., SY/T 5913, SY/T 5368) to ensure scientific rigor and reproducibility. Images were captured using an Axio Scan 7 automated scanning system~\cite{ScanZ2018ZEISSAS}, obtaining multi-channel information (one plane-polarized image and six cross-polarized images) under the same field of view, as shown in Figure \ref{fig: dataset}. These full-frame images were then cropped and sliced into $512 \times 512$ pixel blocks for model training. We avoided direct manual annotation from scratch due to the immense time cost and potential for human error inherent in pixel-level tasks. Instead, we implemented a "\textbf{SAM pre-labeling + expert manual refinement}" strategy. We first utilized the SAM model for automatic rough delineation of grain contours, and then employed professional annotators for strict, pixel-by-pixel review and correction (e.g., ensuring boundary fitting error is under 3 pixels). This process guarantees the high accuracy and standardization of our dataset, providing a robust foundation for advanced intelligent recognition in geological analysis.


\subsection{Petro-SAM Framework Design}
\label{sec: Petro-SAM Framework Design}

\begin{figure}[t]
\centering
\includegraphics[width=\linewidth]{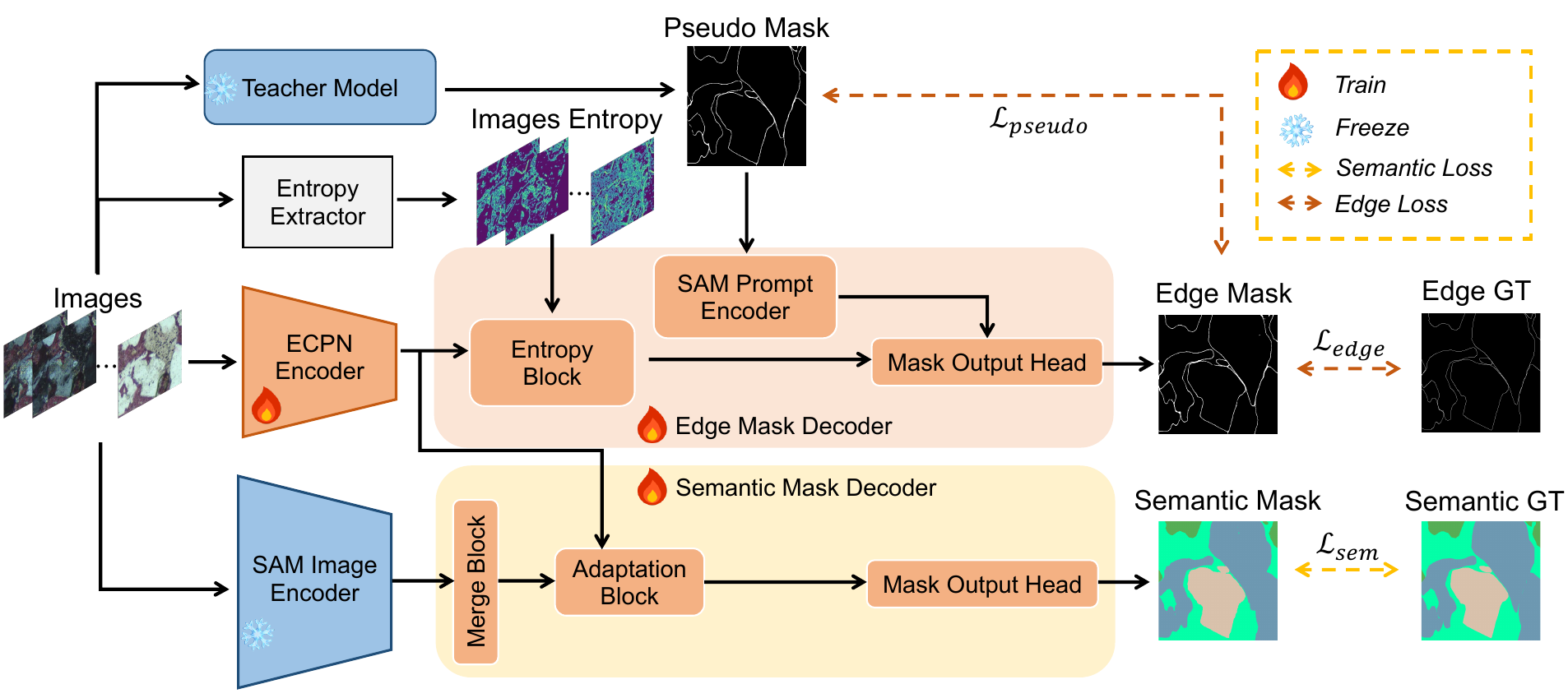}
\caption{Visualization of Petro-SAM's Stage 2 framework}\
\label{fig: stage_2}
\vspace{-0.5cm}
\end{figure}

In this section, we discuss the proposed network Petro-SAM for rock edge segmentation and lithology identification of petrographic images. The construction of Petro-SAM can be divided into two stages. For the first stage, given a group of multi-angled rock petrographic images $X\in \mathbb{R}^{7 \times H \times W \times 3}$ (7 is the number of images per group), we train a model $P_T$ based on SAM's vision transformer encoder \cite{dosovitskiy2020image}, as visualized in Figure \ref{fig: stage_1}, and it outputs an edge mask logit $M_T \in \mathbb{R}^{H \times W}$. Another model $P_S$, visualized in Figure \ref{fig: stage_2} for both edge detection and lithology semantic segmentation, is trained in the second stage. The branch for edge detection takes $M_T$ to prompt feature and output refined edge mask logit $M_S \in \mathbb{R}^{H \times W}$, supervised by both $M_T$ and ground truth $M_G$. In the semantic segmentation branch, $P_S$ integrates the outputs of the SAM image encoder and the encoder of the edge detection branch to provide a semantic mask $Y_S \in \mathbb{R}^{H \times W \times C}$ ($C$ for the number of lithology classes.
We outline the pipeline of the first stage for Petro-SAM in Section \ref{stage 1}. Then, the second stage is detailed in Section \ref{stage 2}. Lastly, in Section \ref{optimization_objective}, we discuss the optimization objective for two stages.

\subsubsection{Stage 1: Teacher Model for Edge Detection}
\label{stage 1}
In stage 1, $X$ is first input into the SAM encoder, which consists of a patch embedding layer, several transformer blocks, convolution layers, and norm layers. We take the shallow feature $\mathcal{S}_{s}\in\mathbb{R}^{7 \times H_e \times W_e \times C_{s}}$ after position embedding addition and deep feature $\mathcal{S}_{d}\in\mathbb{R}^{7 \times H_e \times W_e \times C_{d}}$, where $(H_e, W_e)$ is the shape of the feature map and $C_{s}$ and $C_{d}$ are the number of channels for shallow feature and deep feature, respectively. $\mathcal{S}_{s}$ contains low-level information of simple edge patterns and $\mathcal{S}_{d}$ captures high-level semantic information about entity relationships. Merge Block is designed to combine the features from images of different angles. At the same time, we take the shallow feature $\mathcal{E}_{s}$ and deep feature $\mathcal{E}_{d}$ of the encoder of ECPN \cite{zhang2024edge}. Adaption Block transports $\mathcal{S}_{*}$ from the SAM embedding space to ECPN's. Refine Block refines the deep feature based on the shallow one to extract information related to edge localization. The mask decoder takes the result of the Refine block to output the edge mask logit $M_{T}$.

\noindent \textbf{Merge Block.} Merge Block, which is shown in Figure \ref{fig: stage_1}, is designed to combine the SAM feature of images from different angles. Given feature $\mathcal{S}_{*}\in\mathbb{R}^{7 \times H_e \times W_e \times C_{*}}$, it first extracts the global embedding through global average pooling and compresses the number of channels to 1 through $1\times1$ convolution to obtain image-level weight $\mathcal{W}\in\mathbb{R}^{7}$. $\mathcal{W}$ measures the importance of each image through global embedding. Then the integration of features from different angles can be formulated as:
\begin{align}
\mathcal{S'}_{*}=\frac{\sum_{n=1}^{7} e^{\mathcal{W}[n]} \mathcal{S}_{*}[n]}{\sum_{n=1}^{7} e^{\mathcal{W}[n]}}\in \mathbb{R}^{H_e \times W_e \times C_{*}}
\end{align}

\noindent \textbf{Adaptation Block.} It adapts the SAM feature to the ECPN feature space, as shown in Figure \ref{fig: stage_1}. It first employs adaptive average pooling to align the feature map shape of $\mathcal{S'}_{*}$ to $\mathcal{E}_{*}$. Furthermore, $1\times1$ convolution is utilized to extract channel-wise relationships, and $3\times3$ convolution captures spatial patterns. The concatenated output $\mathcal{S''}_{*}$ is provided to another $1\times1$ convolution to align the channel number to that of $\mathcal{E}_{*}$. At last, Adaption Block attends ECPN feature $\mathcal{E}_{*}$ to $\mathcal{S''}_{*}$ and performs residual addition as follows:
\begin{align}
\mathcal{E'}_{*}=\mathcal{E}_{*} \oplus  Atten(\mathcal{E}_{*}, \mathcal{S''}_{*}, \mathcal{S''}_{*})
\end{align}

\noindent \textbf{Refine Block.} Refine Block refines the deep-layer feature based on that of the shallow-layer. The shape $\mathcal{E'}_{s}$ is aligned with $\mathcal{E'}_{d}$ through adaptive average pooling. Then, the concatenation and $1\times1$ convolution are operated to integrate shallow and deep information, which is followed by element-multiplication to outstand edge-related feature. The process is formulated in Equation \ref{equ: refine block}.
\begin{align}
\mathcal{E''}_{d}=\mathcal{E'}_{d} \otimes Conv_{1\times 1}(Concat(\mathcal{E'}_{s}, \mathcal{E'}_{d}))
\label{equ: refine block}
\end{align}

\subsubsection{Stage 2: Multi-Task Petro-SAM}
\label{stage 2}
In Stage 2, our model further enhances its capability to analyze thin-section images of rocks. As shown in Figure \ref{fig: stage_2}, the model processes multi-angled thin-section images as input and not only outputs grain contour masks but also generates lithology identification masks. 

\begin{algorithm}
\caption{Image Entropy Map Extraction Algorithm}
\label{alg: entropy_map}
\begin{algorithmic}[1]
\REQUIRE A thin-section image $X_i\in\mathbb{R}^{H \times W \times 3}$ of rock and a compression factor $\tau \in \{x|1\leq x\leq7\ \text{and}\ x\in\mathbb{N^+}\}$ for bit-shifting to quantize colors.
\ENSURE Color entropy map $E_{i}\in\mathbb{R}^{H \times W }$.
\STATE Perform bit-shifting by $\tau$ on each color channel of $X_i$ to reduce color resolution.

\STATE Combine RGB channels into a single scalar quantized representation: $$C_i=256\times X_{i}[R]+ \frac{256}{2^\tau} \times X_{i}[G]+X_{i}[B]$$, where $X_{i}[*]$ marks the corresponding color channel.

\STATE Count the number of unique values in $C_i$, denote as $c$. 
\STATE Convert $C_i$ into one-hot encoding color map $C_{i,o}\in \{0, 1\}^{H \times W \times c}$.
\STATE Initiate all-one area matrix $A\in{1}^{H\times W \times 1}$, color kernel $k_{c}\in{1}^{7\times7\times c}$, and area kernel  $k_{a}\in{1}^{7\times7\times 1}$.

\STATE Construct unique color count $C_{i,c}$ by convoluting over $C_{i,o}$ by kernel $k_{c}$ within $c$ groups.

\STATE Construct the area matrix $A_{c}$ of receptive fields by by convoluting over $A$ by kernel $k_{a}$.

\STATE Obtain color entropy map $E_{i}$ through:
$$E_{i}[x,y]=-\sum_{n=1}^{7}\frac{C_{i,c}[x,y,n]}{A_{c}[x, y]}\log\frac{C_{i,c}[x,y,n]}{A_{c}[x, y]}$$

\end{algorithmic}
\end{algorithm}

\begin{figure}[t]
\centering
\includegraphics[width=0.3\linewidth]{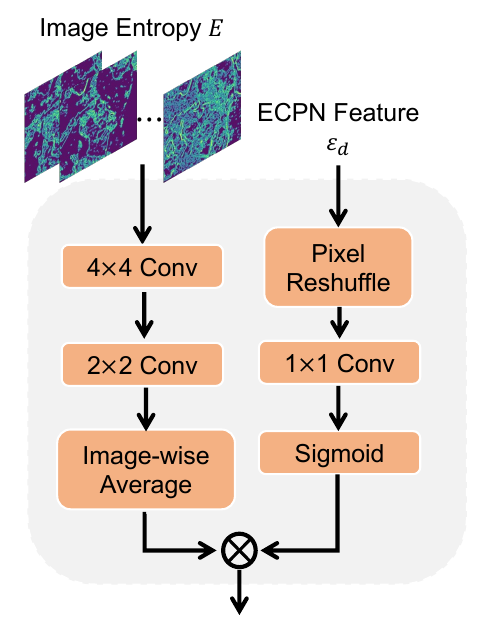}
\caption{Visualization of Entropy Block's structure.}\
\label{fig: entropy_block}
\vspace{-0.5cm}
\end{figure}

\noindent \textbf{Edge Detection Branch.} For the edge extraction task, given the input multi-angled rock petrographic $X$, Petro-SAM leverages the proportion of different colors surrounding each pixel to construct a measure of color entropy $E\in\mathbb{R}^{7\times H \times W \times 3}$, illustrated in Algorithm \ref{alg: entropy_map}. This approach incorporates the prior knowledge embedded in the color entropy, which reflects the boundaries present in the image. By integrating this prior color information, the model enhances its ability to accurately extract contours and delineate grain boundaries in complex thin-section images. Entropy Block, visualized in Figure \ref{fig: entropy_block}, integrates the features extracted by the ECPN encoder with the color entropy information. It achieves this by performing pixel-level feature reorganization to ensure dimensional alignment, followed by the employment of convolutional layers to summarize the color entropy information. Finally, feature refinement is conducted through pixel-wise multiplication, effectively enhancing the representation of relevant features. Next, following the design of SAM, the edge mask logit $M_T$ from the first stage teacher model $P_T$ is used as a prompt to refine and adjust the contours, ensuring the accuracy and clarity of the boundaries. The architecture of the prompt encoder and mask output head is the same as SAM's.

\noindent \textbf{Lithology Segmentation Branch.} For the lithology segmentation task, Petro-SAM adopts a Stage 1-like Merge Block to integrate multi-angled image features from the SAM vision transformer encoder. Adaptation Block fuses outputs from multiple encoders, enabling the decoder to produce lithology recognition results. This process leverages multi-dimensional image data and color features for accurate mineral composition differentiation.

\subsection{Optimization Objectives}
\label{optimization_objective}
For the first stage, given the ground truth edge mask $M_G \in \{0, 1\}^{H \times W}$ and model output logit $M_T$, loss function $\mathcal{L}_{edge,1}$ consists of a class-balanced cross-entropy loss function \cite{xsoria2020dexined, Deng2020DeepSC} $\mathcal{L}_{edge,c}$ and distance map penalized compound loss $\mathcal{L}_{edge,d}$. $\mathcal{L}_{edge,c}$ can be formulated as
\begin{align}
\mathcal{L}_{edge,c}(M_T, M_G)=-\beta_{1}\sum_{M_G[x,y]=1}\log{(M_T[x,y])}-\beta_{0}\sum_{M_G[x,y]=0}\log{(1-M_T[x,y]})
\end{align}
where $\beta_{0}$ and $\beta_{1}$ are the ratio of positive and negative samples in $M_G[x,y]$, respectively. $\mathcal{L}_{edge,d}$ introduces a penalty for the distance to the edge localization to direct the model to concentrate on the rock's boundaries. It can be formulated as
\begin{align}
\mathcal{L}_{edge,d}(M_T, M_G)=-\sum_{(x,y)}\frac{1}{D[x,y]+\epsilon}\log{(M_T[x,y])}
\label{equ: edge_distance}
\end{align}
where $\epsilon$ is the parameter to prevent zero division and $D$ is the distance map that measures the distance of a pixel to the nearest boundary. $\mathcal{L}_{edge,1}$ is the direct summation of $\mathcal{L}_{edge,c}$ and $\mathcal{L}_{edge,d}$.

The second stage is designed for the edge segmentation task and the semantic segmentation task. For the former, the model output $M_S$ is supervised by both $M_T$ and $M_G$. The loss function $\mathcal{L}_{edge,2}$ is shown in Equation \ref{equ: edge_loss_2} where $\lambda_T$ is an adaptation factor that decays with the training process.
\begin{align}
\mathcal{L}_{edge,2}=&(\mathcal{L}_{edge,c}(M_S, M_G) +\mathcal{L}_{edge,d}(M_S, M_G)) +\\
&\lambda_T(\mathcal{L}_{edge,c}(M_S, M_T) +\mathcal{L}_{edge,d}(M_S, M_T))
\label{equ: edge_loss_2}
\end{align}
For the latter, based on Petro-SAM's semantic branch output $M_S$ and ground truth $Y_G\in \{0, 1\}^{H \times W \times C}$, the loss function of semantic segmentation $\mathcal{L}_{sem}$ is the summation of cross entorpy loss $\mathcal{L}_{sem,ce}$ and dice loss $\mathcal{L}_{sem,d}$, defined as 
\begin{align}
\mathcal{L}_{sem,ce}=-\frac{1}{HW}\sum_{(x,y)}\sum_{c=1}^{C}Y_G[x,y,c]\log(M_S[x,y,c])
\\
\mathcal{L}_{sem,d}=1-\frac{1}{C}\sum_{c=1}^{C}\frac{2\sum_{(x,y)}Y_G[x,y,c]M_S[x,y,c]}{\sum_{(x,y)}(Y_G[x,y,c]^2+M_S[x,y,c]^2+\epsilon)}
\end{align}
where $\epsilon$ is the same parameter as in Equation \ref{equ: edge_distance}. This optimization strategy significantly improves the accuracy and stability of composition recognition, ensuring the model performs exceptionally well in the analysis of complex rock images. The total loss for stage 2 model training is formulated in Equation \ref{equ: stage_2_loss}, where $\lambda_{e}$ is the balancing factor.
\begin{align}
\mathcal{L}=\lambda_{e}\mathcal{L}_{edge, 2}+\mathcal{L}_{sem}
\label{equ: stage_2_loss}
\end{align}














\section{Experiment}
\label{Experiment}
\subsection{Experimental Setup}
\noindent \textbf{Dataset.} Our experiments were conducted using two complementary datasets: the public CPPID dataset and the constructed petrographic thin-section dataset. The CPPID dataset contains 30,625 images grouped into 4,375 paired samples, with 3,500 groups allocated for training and 875 for testing. Each image is provided with pixel-level edge annotations, enabling supervised learning of fine-grained boundary structures across diverse mineral textures. In addition, we employed our dataset comprising 1,400 image groups acquired under consistent petrographic imaging conditions. Among these, 1,120 groups were used for training and 280 for testing. This dataset includes not only pixel-level edge labels but also detailed semantic annotations. The semantic categories cover quartz, feldspar, debris, and background, allowing joint learning of edge detection and multi-class semantic segmentation. The complementary characteristics of the two datasets—large-scale edge supervision from CPPID and rich mineralogical semantics from our dataset—provide a comprehensive benchmark for evaluating the effectiveness and robustness of our proposed segmentation framework.

\noindent \textbf{Implementation Details.} Petro-SAM was implemented using the PyTorch framework. The encoder integrates two sources of pretrained weights: the ViT \cite{dosovitskiy2020image} backbone from the SAM \cite{kirillov2023segment} and an EfficientNetV2 \cite{Tan2021EfficientNetV2SM} backbone, both of which provide strong initialization for handling complex petrographic textures. We adopted the Adam optimizer with $ \beta=(0.9, 0.999)$, $\epsilon = 1e-8$, and a weight decay of $2e-4$. The initial learning rate was set to $2e-4$, and we applied a cosine annealing warm-restarts scheduler with a minimum learning rate of $2e-6$ to ensure stable convergence during later training phases. The entire training process was divided into two stages. In Stage 1, the model was trained for 50 epochs with a batch size of 4 to learn robust mineral boundaries and semantic structures. Stage 2 further refined the model for 50 epochs using the same batch size, with $\lambda_{e}$ being 0.0004. All experiments were conducted on a single NVIDIA A800-80GB GPU, and the full training pipeline required approximately 20 hours. 

 \begin{table}[t]
\centering
\caption{Comparisons of the boundary detection performance of the proposed Petro-SAM and prior arts.}
\label{tab:Detection_comparison}
\resizebox{\textwidth}{!}{
\begin{tabular}{lllccccccccccc}
\toprule
\multicolumn{2}{l}{\multirow{2}{*}{Method}}               &&\multicolumn{5}{c}{Our Dataset}                   &&\multicolumn{5}{c}{CPPID}      
\\
\cdashline{4-8} \cdashline{10-14}

\multicolumn{2}{l}{}                                      && mIoU   & F1-Score & Precision & Recall & Accuracy && mIoU   & F1-Score & Precision & Recall & Accuracy 
\\ 
\hline

\multicolumn{14}{l}{\cellcolor[HTML]{EFEFEF}\textit{Traditional Methods}} 
\\ 
\hline
\multicolumn{2}{l}{XDoG \cite{10.1145/2024676.2024700}}  && 2.8\%  & 0.053    & 0.034     & 0.212  & 0.962    && 6.5\%  & 0.121    & 0.156     & 0.121  & 0.930    
\\
\multicolumn{2}{l}{SE \cite{dollar2014fast}}             && 1.8\%  & 0.036    & 0.018     & 0.985  & 0.238    && 6.9\%  & 0.128    & 0.069     & 0.986  & 0.340     
\\
\multicolumn{2}{l}{HED \cite{xie2015holistically}}       && 5.5\%  & 0.103    & 0.056     & 0.773  & 0.800    && 16.5\% & 0.282    & 0.173     & 0.795  & 0.796    
\\
\multicolumn{2}{l}{RCF \cite{liu2017richer}}             && 5.3\%  & 0.101    & 0.056     & 0.559  & 0.851    && 16.4\% & 0.281    & 0.170     & 0.823  & 0.786    
\\
\multicolumn{2}{l}{BDCN \cite{he2019bi}}                  && 20.7\% & 0.343    & 0.212     & 0.899  & 0.950    && 33.5\% & 0.501    & 0.340     & 0.956  & 0.904    
\\
\multicolumn{2}{l}{Dexined \cite{xsoria2020dexined}}     && 17.9\% & 0.303    & 0.182     & \textbf{0.915}  & 0.938    && 25.0\% & 0.399    & 0.253     & \textbf{0.959}  & 0.852    
\\
\multicolumn{2}{l}{PiDiNet \cite{su2021pixel}}           && 13.9\% & 0.244    & 0.141     & 0.909  & 0.918    && 20.1\% & 0.333    & 0.205     & 0.908  & 0.816    
\\
\multicolumn{2}{l}{ECPN \cite{zhang2024edge}}            && 22.0\% & 0.360    & 0.238     & 0.741  & 0.961    && 35.3\% & 0.521    & 0.362     & 0.931  & 0.914    
\\ 
\hline
\multicolumn{14}{l}{\cellcolor[HTML]{EFEFEF}\textit{SAM-based Methods}} 
\\ 
\hline

\multirow{3}{*}{SAM \cite{kirillov2023segment}}           & vit-b && 5.3\%  & 0.099    & 0.055     & 0.565  & 0.846    && 11.5\% & 0.205    & 0.160     & 0.328  & 0.880    
\\
& vit-l && 6.9\%  & 0.129    & 0.074     & 0.518  & 0.895    && 12.9\% & 0.226    & 0.199     & 0.290  & 0.907    
\\
 & vit-h && 7.1\%  & 0.131    & 0.075     & 0.535  & 0.893    && 13.7\% & 0.239    & 0.197     & 0.324  & 0.900    
 \\
 
\multirow{2}{*}{SAUGE \cite{liufu2025sauge}}              & vit-b && 6.6\%  & 0.123    & 0.067     & 0.826  & 0.826    && 16.0\% & 0.275    & 0.166     & 0.839  & 0.778    
\\
& vit-l && 6.5\%  & 0.121    & 0.066     & 0.815  & 0.824    && 16.0\% & 0.274    & 0.166     & 0.824  & 0.782    
\\ 
\hline
\multicolumn{14}{l}{\cellcolor[HTML]{EFEFEF}\textit{Our Method}} 
\\ 
\hline

\multirow{3}{*}{Teacher Model} & vit-b && 27.8\% & 0.432    & 0.308     & 0.724  & 0.972    && 41.4\% & 0.584    & 0.427     & 0.927  & 0.934    
\\
& vit-l && 29.7\% & 0.454    & 0.328     & 0.740  & 0.974    && 43.7\% & 0.606    & 0.450     & 0.933  & 0.939    
\\
& vit-h && 31.1\% & 0.471    & 0.344     & 0.753  & 0.975    && 44.2\% & 0.611    & 0.457     & 0.923  & 0.941    
\\

\multirow{3}{*}{Petro-SAM} & vit-b && 42.6\%      & 0.592        & 0.552         & 0.638      & 0.987        && 41.7\%      & 0.589        & 0.430         & 0.936      & 0.938        
\\
 & vit-l && 45.3\%      & 0.624        & 0.573         & 0.684      & 0.989        && 43.8\%      & 0.609        & 0.451         & 0.938      & 0.940        
 \\
& vit-h && \textbf{48.2\%}      & \textbf{0.650}        & \textbf{0.630}        & 0.672       & \textbf{0.990}        && \textbf{44.4\%}      & \textbf{0.615}        & \textbf{0.460}         & 0.928      & \textbf{0.942}      
\\
\bottomrule
\end{tabular}}
\end{table}

\subsection{Boundary Detection Results}
Table \ref{tab:Detection_comparison} details the quantitative comparison of our proposed method against Traditional (e.g., BDCN~\cite{he2019bi}, ECPN~\cite{zhang2024edge}) and SAM-based approaches on Our Dataset and the public CPPID. Our Method, specifically the Petrao-SAM (vit-h) variant, achieves state-of-the-art (SOTA) performance across all key metrics on both datasets. On Our Dataset, Petrao-SAM (vit-h) secured the highest mIoU of 48.2\% and F1-Score of 0.650. This marks a substantial margin over the best traditional method, BDCN (mIoU: 20.7\%), and significantly outperforms SAM-based approaches (e.g., SAM vit-h, mIoU: 7.1\%), underscoring its enhanced capability for dense prediction tasks. Furthermore, to ensure a fair and generalized assessment of our framework on an external, publicly available domain, we also validated our method on the CPPID dataset. The data proves that this effectiveness is maintained, with Petrao-SAM (vit-h) achieving an impressive mIoU of 44.4\% and F1-Score of 0.615, setting a new performance record against the highest competing traditional method, ECPN (mIoU: 35.3\%). These consistent improvements across two diverse datasets affirm that our proposed method provides a powerful and robust solution for accurate segmentation.

\begin{figure}[!t]
\centering
\includegraphics[width=\linewidth]{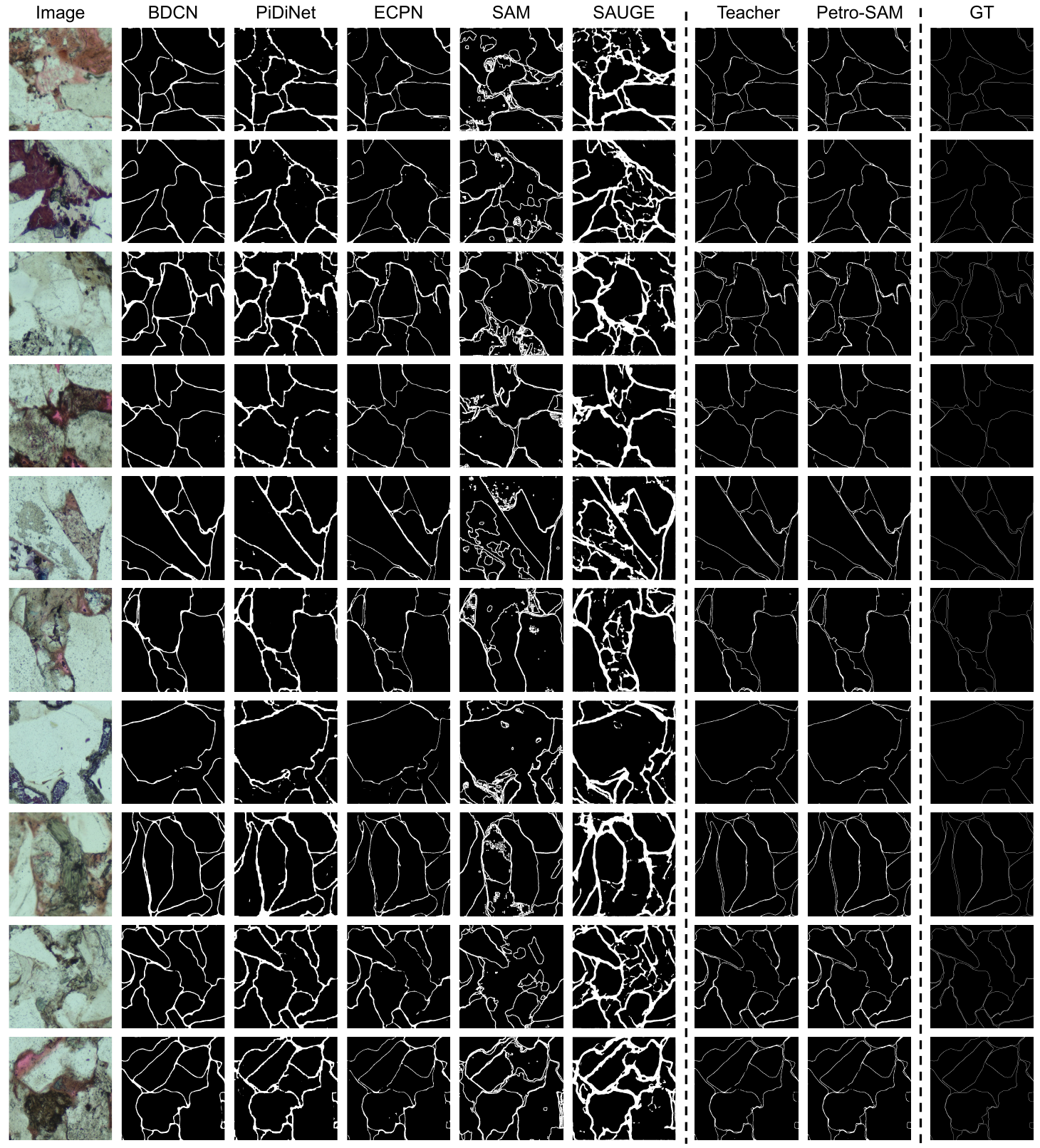}
\caption{Qualitative edge results comparing existing methods with our Petro-SAM.}\
\label{fig: edge_comparision}
\vspace{-0.5cm}
\end{figure}

 Beyond the strong quantitative results, the qualitative visualization in Figure \ref{fig: edge_comparision} further confirms the superior performance and high fidelity of our model. The outputs of traditional edge detectors (BDCN, PiDiNet, ECPN) often suffer from over-segmentation and noise, while the raw SAM and SAUGE models fail dramatically, producing chaotic and low-fidelity boundaries that are inconsistent with the Ground Truth (GT).  In contrast, the edge detection results from our Petrao-SAM model closely resemble the GT, producing clean, continuous, and simplified grain contours that successfully exclude the noise prevalent in competing methods. The alignment between the SOTA metrics and the high-quality visualization across both self-collected and public domains mutually affirms that our proposed Petro-SAM framework provides a robust and reliable solution for accurate geological grain edge segmentation.

\subsection{Semantic Segmentation Results}
\begin{table}[t]
\centering
\caption{Comparisons of the semantic segmentation performance of the proposed Petro-SAM and prior arts.}
\label{tab:Segmentation_comparison}
\resizebox{\textwidth}{!}{
\begin{tabular}{lllccccccc}
\toprule
\multicolumn{2}{l}{Method}           && mIoU   & Background & Feldspar & Debris & Quartz && Accuracy \\ \midrule
\multicolumn{2}{l}{DeepLabV3 \cite{chen2017rethinkingatrousconvolutionsemantic}}        && 83.8\% & 84.3\%     & 75.6\%   & 85.9\% & 89.5\% && 0.900    \\
\multicolumn{2}{l}{CCNet \cite{huang2019ccnet}}            && 83.4\% & 83.7\%     & 75.5\%   & 86.4\% & 88.3\% && 0.897    \\
\multicolumn{2}{l}{FCN \cite{long2015fully}}              && 85.0\% & 87.2\%     & 75.8\%   & 88.2\% & 89.0\% && 0.906    \\
\multicolumn{2}{l}{PSPNet \cite{zhao2017pyramid}}           && 83.1\% & 83.7\%     & 74.3\%   & 85.3\% & 89.1\% && 0.899    \\
\multicolumn{2}{l}{Swin-UNet \cite{cao2022swin}}        && 76.6\% & 75.2\%     & 70.6\%   & 78.7\% & 82.0\% && 0.845    \\ \midrule
\multirow{3}{*}{SAM \cite{kirillov2023segment}}         & vit-b  && 40.6\% & 49.4\%     & 23.5\%   & 32.6\% & 56.8\% && 0.549    \\
                             & vit-l  && 34.9\% & 44.1\%     & 14.6\%   & 31.5\% & 49.4\% && 0.493    \\
                             & vit-h  && 34.1\% & 44.2\%     & 13.6\%   & 29.8\% & 48.9\% && 0.485    \\ \midrule
\multirow{2}{*}{Mask2Former \cite{cheng2021mask2former}} & swin-b && 56.2\% & 60.9\%     & 43.4\%   & 54.2\% & 66.3\% && 0.705    \\
                             & swin-l && 57.4\% & 61.3\%     & 44.2\%   & 57.4\% & 66.5\% && 0.716    \\ \midrule
\multirow{3}{*}{Petro-SAM}   & vit-b  && 83.4\%      & 85.7\%          & 73.3\%        & 85.6\%      & 88.9\%      & &0.900        \\
                             & vit-l  && 84.5\%      & 82.6\%
& 78.0\%        & 85.9\%      & 91.5\%      && 0.911        \\
                             & vit-h  && \textbf{86.3\%}      & \textbf{87.6\%}     
& \textbf{75.9\%}        & \textbf{89.2\%}      & \textbf{92.4\%}      && \textbf{0.928}        \\ \bottomrule
\end{tabular}}
\end{table}

Table \ref{tab:Segmentation_comparison} compares our proposed Petro-SAM method against classic models (e.g., FCN \cite{long2015fully}, DeepLabV3 \cite{chen2017rethinkingatrousconvolutionsemantic}), Mask2Former \cite{cheng2021mask2former}, and the foundational SAM model \cite{kirillov2023segment}. The results clearly demonstrate that our approach achieves state-of-the-art (SOTA) performance across all evaluation metrics. Petro-SAM (vit-h) secured the highest overall performance, achieving an impressive mIoU of 86.3\% and an Accuracy of 0.928. This significantly surpasses the best traditional method, FCN (mIoU 85.0\%). Furthermore, Petro-SAM demonstrates superior segmentation performance in challenging categories, notably achieving an IoU of 75.9\% for Feldspar. The dramatic improvement over the raw SAM (vit-b), whose mIoU was only 40.6\% compared to our Petro-SAM (vit-b) at 83.4\%, strongly validates the effectiveness and robustness of our proposed architecture and adaptation strategy in successfully applying SAM's knowledge to the complex domain of mineral segmentation.

\begin{figure}[!t]
\centering
\includegraphics[width=\linewidth]{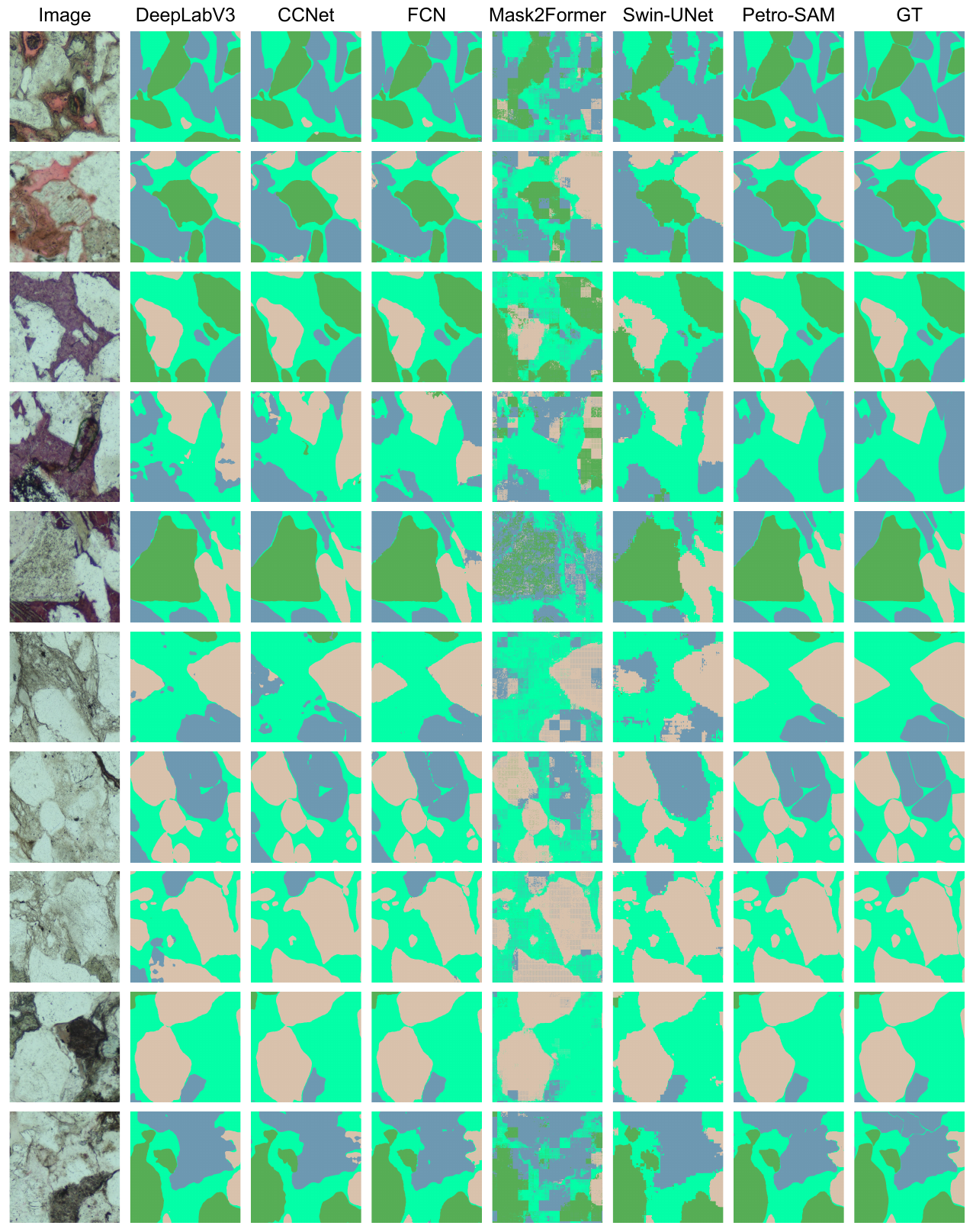}
\caption{Qualitative semantic results comparing existing methods with our Petro-SAM.}
\label{fig: semantic_comparison}
\vspace{-0.5cm}
\end{figure}

Beyond the strong quantitative results, the qualitative visualization in Figure \ref{fig: semantic_comparison} Semantic Comparison further validates the high fidelity of our model in segmenting mineral categories. While classic models (FCN, DeepLabV3) and Swin-UNet often produce masks with jagged boundaries and localized category confusion, and Mask2Former exhibits fragmentation and irregular holes (e.g., rows 4 and 5), our Petro-SAM model consistently generates results that are closest to the GT. Petro-SAM successfully produces clean, smooth, and regionally continuous segmentation masks across all samples. This demonstrates that our framework not only achieves superior mIoU but also excels in accurately localizing regional boundaries and maintaining high semantic consistency, confirming its reliability for accurate lithological recognition in complex geological images.

\subsection{Ablation Study of Stage 1}
\label{Ablation Study of Stage 1}
\begin{table}[]
\centering
\label{tab: stage_1_ablation}
\caption{Ablation results for the three components of our Stage 1 teacher model.}
\resizebox{0.6\textwidth}{!}{
\begin{tabular}{cccccc}
\toprule
Merge Block & Adaption Block & Refine Block && mIoU  & F-Score \\ \midrule
            &                &              && 22.0\% & 0.360   \\
\ding{51}         &                &              && 24.0\%   & 0.384     \\
            & \ding{51}            &              && 23.7\% & 0.380   \\
            &                & \ding{51}          && 22.2\%   & 0.361     \\
\ding{51}         & \ding{51}            &             && 27.1\% & 0.423   \\
            & \ding{51}            & \ding{51}         && 23.8\% & 0.382   \\
\ding{51}         &                & \ding{51}         && 25.0\% & 0.383   \\
\ding{51}         & \ding{51}            & \ding{51}   && 27.8\% & 0.432   \\ 
\bottomrule
\end{tabular}}
\end{table}

\begin{figure}[t]
\centering
\includegraphics[width=\linewidth]{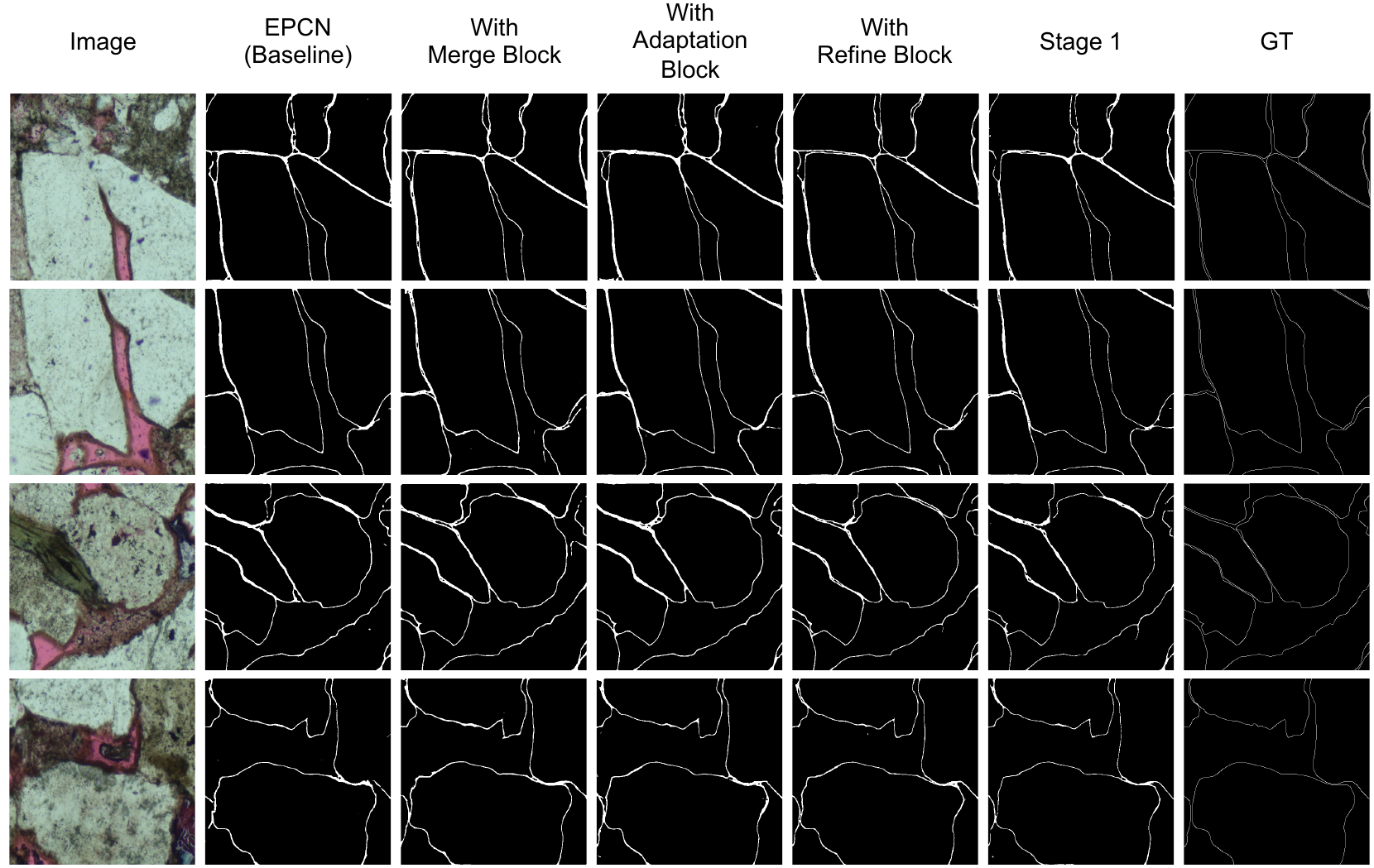}
\caption{Stage 1 Ablation Visualization. Zoom in for a better understanding.}\
\label{fig: stage_1_ablation}
\vspace{-0.5cm}
\end{figure}

In Stage~1, we designed a dual-branch teacher model to exploit the complementary strengths of SAM’s high-level structural priors and ECPN’s extinction-aware representations. To validate the effectiveness of this design, we performed a comprehensive ablation study on the three core components of Stage~1: the Merge Block, Adaptation Block, and Refine Block (Table \ref{tab: stage_1_ablation}). The dual-branch design was motivated by the observation that SAM’s encoder, although robust to texture variability, lacked the petrographic sensitivity required to capture extinction-driven grain boundaries, while the ECPN encoder captured extinction cues but produced fragmented or noisy contours without global contextual regularization. The Merge Block addressed this gap by fusing seven polarized views using global-attention pooling and image-wise summation, yielding extinction-consistent features and improving mIoU from 22.0\% to 24.0\%. The Adaptation Block further aligned SAM’s feature space with petrographic textures through attention-based feature transformation, raising performance to 23.7\% mIoU. The Refine Block enhanced the delineation of thin and low-contrast grain edges through multi-scale aggregation, offering moderate improvements when used alone. When the blocks were progressively combined, performance increased steadily, with the full Stage~1 model achieving the best results at 27.8\% mIoU and 0.432 F-score. These results demonstrated that multi-angle fusion, cross-domain adaptation, and fine-grained refinement were all necessary and complementary. As shown in our qualitative comparisons (Figures \ref{fig: edge_comparision} and \ref{fig: stage_1_ablation}), the Stage~1 teacher generated extinction-aware edge maps that were significantly cleaner and sharper than those produced by classical edge detectors or SAM alone. These high-quality edge prompts subsequently enabled Stage~2 to perform more accurate GES–LSS joint learning, confirming the essential role of Stage~1 in the overall Petro-SAM framework.

\subsection{Ablation Study of Stage 2}
\label{Ablation Study of Stage 2}
\begin{table}[]
\centering
\caption{Ablation results for Stage 2 edge branch.}
\label{tab: stage_2_EDGE_ablation}
\resizebox{0.6\textwidth}{!}{
\begin{tabular}{ccccccc}
\toprule
$\mathcal{L}_{sem}$ & $\mathcal{L}_{edge}$ & Entropy Block & Stage 1 && mIoU  & F-Score \\ \midrule
 & \ding{51} & \ding{51} & \ding{51} && 43.2\% & 0.604   \\
\ding{51} & & \ding{51} & \ding{51} && 27.7\% & 0.431   \\
\ding{51} & \ding{51} & & \ding{51} && 38.3\% & 0.549   \\
\ding{51} & \ding{51} & \ding{51} & && 34.7\% & 0.506   \\
\ding{51} & \ding{51} & \ding{51} & \ding{51}   && 42.6\% & 0.592   \\ 
\bottomrule
\end{tabular}}
\end{table}

\begin{table}[]
\centering
\caption{Ablation results for Stage 2 semantic branch.}
\label{tab: stage_2_sem_ablation}
\resizebox{0.6\textwidth}{!}{
\begin{tabular}{ccccc}
\toprule
$\mathcal{L}_{edge}$ & Adaptation Block && mIoU  & Accuracy \\ \midrule
 \ding{51} & && 59.4\% & 0.728   \\
  & \ding{51} && 75.3\% & 0.835   \\
\ding{51} & \ding{51} && 83.4\% & 0.900   \\
\bottomrule
\end{tabular}}
\end{table}

\begin{figure}[t]
\centering
\includegraphics[width=\linewidth]{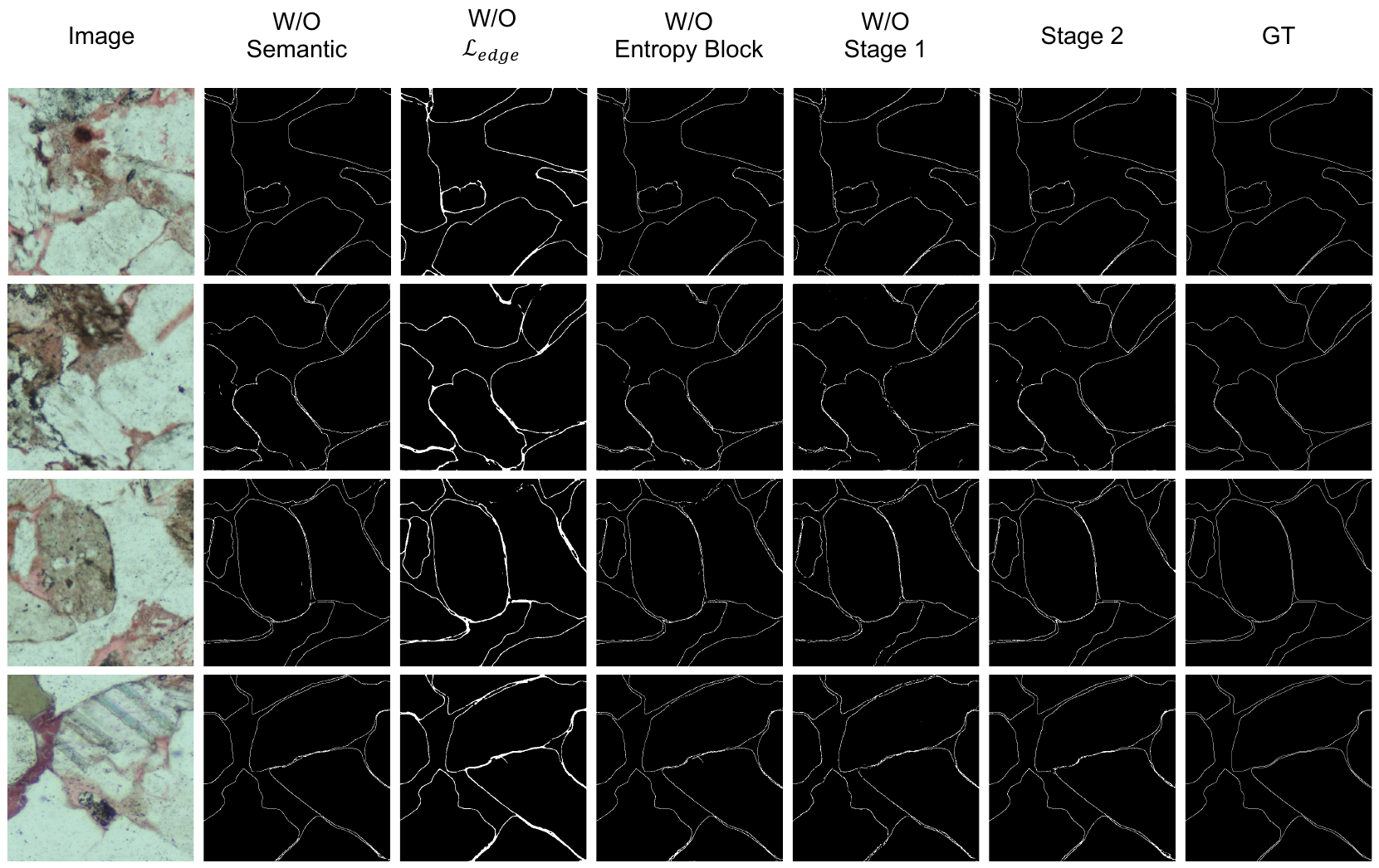}
\caption{Stage 2 Edge Ablation Visualization. Zoom in for a better understanding.}\
\label{fig: stage_2_edge_ablation}
\vspace{-0.5cm}
\end{figure}

\noindent \textbf{Ablation study of edge branch.}
In Stage~2, we investigated the behavior of the edge branch under multi-task learning, where both the edge and semantic heads share the same ECPN encoder. Table \ref{tab: stage_2_EDGE_ablation} presents a leave-one-out ablation in which each component of the Stage~2 edge branch is individually removed and compared to the full model (42.6\% mIoU, 0.592 F-score). 
This setup allows us to quantify the specific contribution of each element under multi-task learning, where the edge and semantic heads share the same ECPN encoder.
When semantic supervision was excluded, the model reached 43.2\% mIoU and a 0.604 F-score, slightly higher than the full configuration. 
This indicates that $\mathcal{L}_{sem}$ introduces moderate gradient competition in the shared encoder; however, its inclusion in the full model remains necessary for producing coherent lithologic boundaries, even if it mildly constrains edge sharpness.
Eliminating the edge loss caused the strongest degradation, reducing performance to 27.7\% mIoU and 0.431 F-score. 
The comparison with the full model highlights that $\mathcal{L}_{edge}$ is essential for supervising fine-grained contour structures and preventing the encoder from being dominated by semantic gradients.
Disabling the entropy block reduced accuracy to 38.3\% mIoU and 0.549 F-score. 
Relative to the full model, this confirms that entropy-guided modulation stabilizes extinction-dependent or low-contrast boundaries, preventing them from being suppressed by multi-task gradients.
Removing Stage~1 prompts further lowered performance to 34.7\% mIoU and 0.506 F-score. 
As visualized in Figure \ref{fig: stage_2_edge_ablation}, the contrast with the full model demonstrates that the extinction-consistent priors transferred from Stage~1 provide crucial structural guidance, anchoring the edge branch during joint optimization.
Overall, the full configuration delivered the best balance between edge precision and semantic-phase consistency. 
Although multi-task supervision introduces mild competition, combining all components ultimately produced the most coherent and extinction-consistent edge maps, validating the necessity of tightly coupling edge and semantic learning in Petro-SAM.

\noindent \textbf{Ablation study of semantic branch.} Table \ref{tab: stage_2_sem_ablation} evaluates how the semantic Adaptation Block and the edge loss $\mathcal{L}_{edge}$ contribute to lithology segmentation by progressively enabling each component.
With only $\mathcal{L}_{edge}$ active and the Adaptation Block removed, the semantic head reached 59.4\% mIoU and 0.728 accuracy. 
In this setting, the edge branch regularized the shared encoder with boundary-sensitive cues; however, without feature alignment from the Adaptation Block, the semantic decoder lacked sufficient high-level semantic structure, resulting in over-smoothed predictions and class mixing across grains.
When we disabled $\mathcal{L}_{edge}$ but enabled the Adaptation Block, performance improved substantially to 75.3\% mIoU and 0.835 accuracy. 
This confirms that injecting SAM-derived semantic structure into the shared encoder provides strong global context for lithologic discrimination. 
Nevertheless, in the absence of edge supervision, the model failed to preserve fine-grain contacts, causing neighboring grains of similar appearance to remain partially blended.
The full configuration, which activates both the Adaptation Block and $\mathcal{L}_{edge}$, achieved the best results at 83.4\% mIoU and 0.900 accuracy. 
\begin{figure}[t]
\centering
\includegraphics[width=0.7\linewidth]{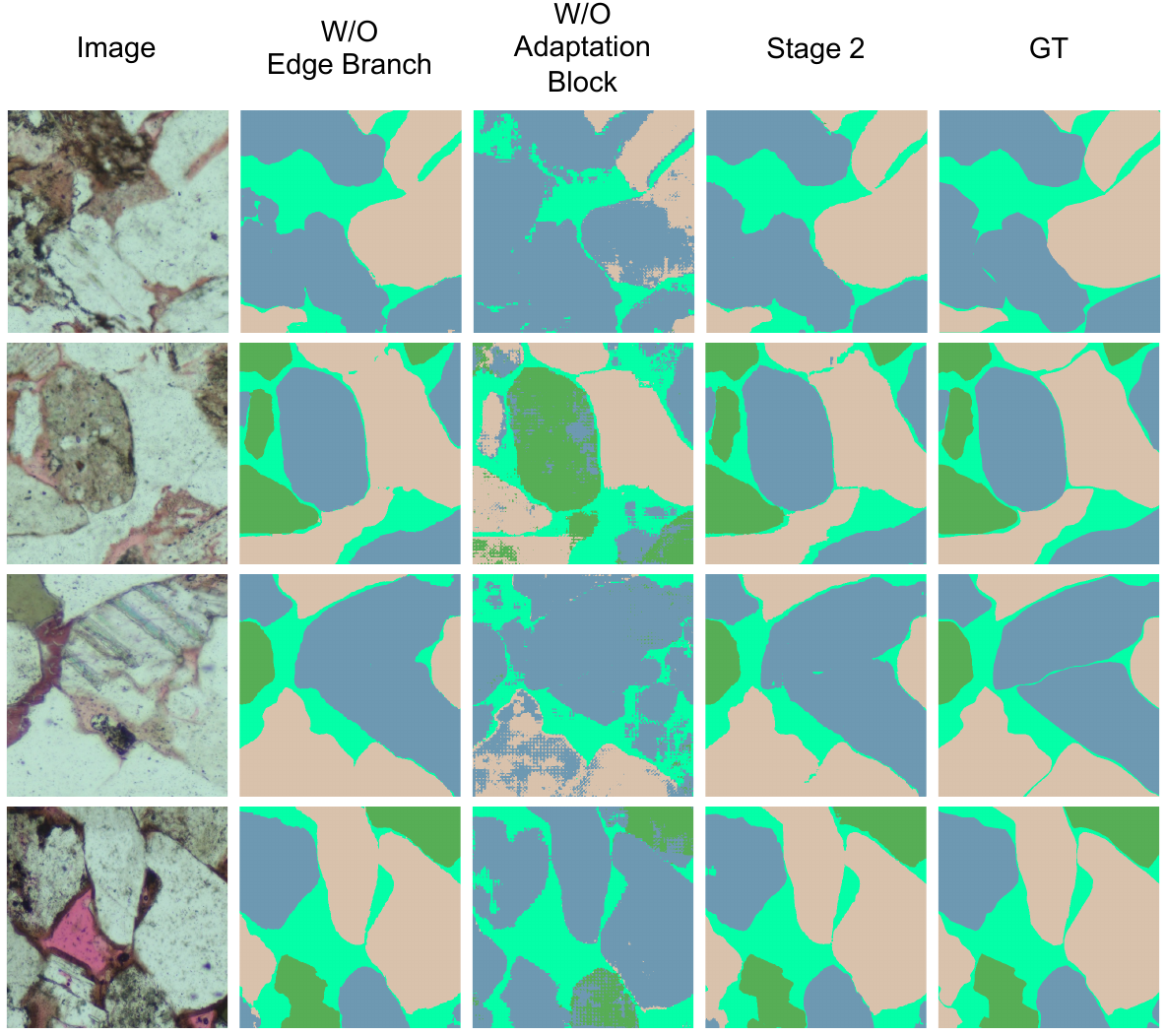}
\caption{Stage 2 Semantic Ablation Visualization. Zoom in for a better understanding.}\
\label{fig: stage_2_semantic_ablation}
\vspace{-0.5cm}
\end{figure}
Here, the semantic pathway benefited simultaneously from SAM-informed semantic alignment and explicit boundary cues propagated through the shared encoder, enabling the decoder to maintain interior consistency while accurately separating adjacent mineral grains. 
As visualized in Figure \ref{fig: stage_2_semantic_ablation}, the full model produced coherent mineral-phase regions with clearly preserved grain boundaries, whereas models lacking either component exhibited region bleeding or boundary artifacts.
These findings demonstrate that the Adaptation Block and edge-guided supervision are complementary: the former provides global semantic structure, while the latter enforces local boundary precision. Their combination is essential for high-quality lithology segmentation in Petro-SAM.

\section{Conclusion}
\label{Conclusion}

In this work, we introduced Petro-SAM, a two-stage framework for automated petrographic thin-section understanding under multi-angle polarized imaging. We also constructed a new multi-angle polarized thin-section dataset, comprising seven-view polarized stacks with expert-annotated grain-edge masks and four-class lithology labels, providing the first benchmark that simultaneously supports both grain-edge segmentation (GES) and lithology semantic segmentation (LSS).
Building upon this dataset, our Stage-1 teacher model fused polarized views to extract extinction-consistent boundary cues, adapted SAM’s encoder features to petrographic textures, and generated high-fidelity edge prompts. Stage-2 jointly learned GES and LSS through a multi-task student network, integrating prompt-guided boundary refinement, a color-entropy prior, and boundary-aware losses to enhance contour sharpness and mineral-phase separability.
Comprehensive experiments demonstrated that Petro-SAM consistently outperformed classical edge detectors, extinction-based networks, promptable segmentation models, and strong semantic baselines. Our model achieved clearer grain boundaries, more coherent lithologic regions, and superior robustness to polarization-induced variations. Overall, this study showed that combining multi-angle cues, two-stage prompt-guided learning, and joint GES–LSS supervision offered a principled path forward for accurate and scalable digital petrography.

\section*{Acknowledgement}
This work was financially supported by the National Natural Science Foundation of China (No. 42372175), National Science and Technology Major Project-Key Equipment and Software for Smart Oil and Gas Fields (No. 2025ZD1401501).

\section*{Declaration of competing interests}
The authors declare that they have no known competing financial interests or personal relationships that could have appeared to influence the work
reported in this paper.

\section*{Declaration of generative AI and AI-assisted technologies in the writing process}
During the preparation of this work the author(s) used [ChatGPT] in order to [check
grammar and improve language]. After using this tool/service, the author(s) reviewed
and edited the content as needed and take(s) full responsibility for the content of the
publication.






\clearpage
\bibliographystyle{IEEEtran}
\bibliography{main}



\end{document}